\documentclass[letterpaper, 10 pt, conference]{ieeeconf}  % Comment this line out if you need a4paper

\IEEEoverridecommandlockouts                              % This command is only needed if 
\usepackage{graphics} % for pdf, bitmapped graphics files
\usepackage{epsfig} % for postscript graphics files
\usepackage{mathptmx} % assumes new font selection scheme installed
\usepackage{times} % assumes new font selection scheme installed
\usepackage{amsmath} % assumes amsmath package installed
\usepackage{amssymb}  % assumes amsmath package installed
\usepackage{amsfonts}
\usepackage{hyperref}
\usepackage{verbatim}
\usepackage{subfigure}
\usepackage{multirow}
\usepackage{booktabs}
\usepackage{wrapfig}
\usepackage[english]{babel}
\usepackage{algorithm}
\usepackage{algorithmic}
\usepackage{xcolor}

\newcommand{\TP}{\textit{TP}}

\newcommand{\EGAE}{\text{EGAE}}
\newcommand{\raw}{\textit{raw}}

\title{\LARGE \bf
	Reinforcement Learning for Robot Navigation with Adaptive Forward
	Simulation Time (AFST) in a Semi-Markov Model
}

\author{Yu'an Chen, Ruosong Ye, Ziyang Tao, Hongjian Liu, Guangda Chen, Jie Peng, Jun Ma, \\ Yu Zhang, Jianmin Ji$^*$, and Yanyong Zhang$^*$% <-this % stops a space
	\thanks{School of Computer Science and Technology, University of Science and Technology of China, Hefei, 230026, China}%
	\thanks{$^*$ Corresponding authors {\tt\small \{jianmin,yanyongz\}@ustc.edu.cn}}%
	\thanks{The work is partially supported by Guangdong Province R\&D Program 2020B0909050001, Anhui Province Development and Reform Commission 2020 New Energy Vehicle Industry Innovation Development Project and 2021 New Energy and Intelligent Connected Vehicle Innovation Project, and Shenzhen Yijiahe Technology R\&D Co., Ltd.}
}

\begin{document}
	
	\maketitle
	\thispagestyle{empty}
	\pagestyle{empty}
	
	%%%%%%%%%%%%%%%%%%%%%%%%%%%%%%%%%%%%%%%%%%%%%%%%%%%%%%%%%%%%%%%%%%%%%%%%%%%%%%%%
	\begin{abstract}
		Deep reinforcement learning (DRL) algorithms have proven effective in robot navigation, especially in unknown environments, by directly mapping perception inputs into robot control commands. 
		However, most existing methods ignore the local minimum problem in navigation and thereby cannot handle complex unknown environments. 
		In this paper, we propose the first DRL-based navigation method modeled by a semi-Markov decision process (SMDP) with continuous action space, named Adaptive Forward Simulation Time (AFST), to overcome this problem. Specifically, we reduce the dimensions of the action space and improve the distributed proximal policy optimization (DPPO) algorithm for the specified SMDP problem by modifying its GAE to better estimate the policy gradient in SMDPs. 
		Experiments in various unknown environments demonstrate the effectiveness of AFST.
	\end{abstract}

	\section{Introduction}\label{intro}
	Autonomous robot navigation with complete maps has been extensively studied in recent years~\cite{lavalle2006planning}. 
	However, in many real-world scenarios such as search and rescue operations~\cite{niroui2019deep} or rapidly changing environments~\cite{marchesini2021benchmarking}, global maps are not always available.
	Conventional path-planning approaches~\cite{Astar,rrt} heavily rely on accurate, static models of the environment, making them ineffective when such maps are unknown.
	Consequently, there is a need for robust navigation approaches that can help robots navigate in unknown environments using local perception.
	
	In recent years, Deep reinforcement learning (DRL) algorithms have shown promising results in navigation tasks in unknown environments~\cite{tai2017virtual,ral2018mapless,icra2020discrete}. 
	These data-driven methods learn navigation policies through trial-and-error exploration combined with carefully designed reward signals. Typically, these policies directly map perception inputs, such as 2D laser data or local costmaps, to low-level control commands like translational and rotational velocities. 
	However, existing methods often overlook the local minima problem~\footnote{The local minima problem in robot navigation usually refers to a physical situation where a robot gets trapped in a suboptimal state while trying to navigate from its current location to a desired goal~\cite{lmp}.}, which is more common and challenging in unknown environments. As a result, they can only handle relatively simple environments without dead ends. This paper investigates the use of DRL to address this problem by formulating the navigation task differently.
	
\begin{figure}[tp]
		\centerline{\includegraphics[width=1\linewidth]{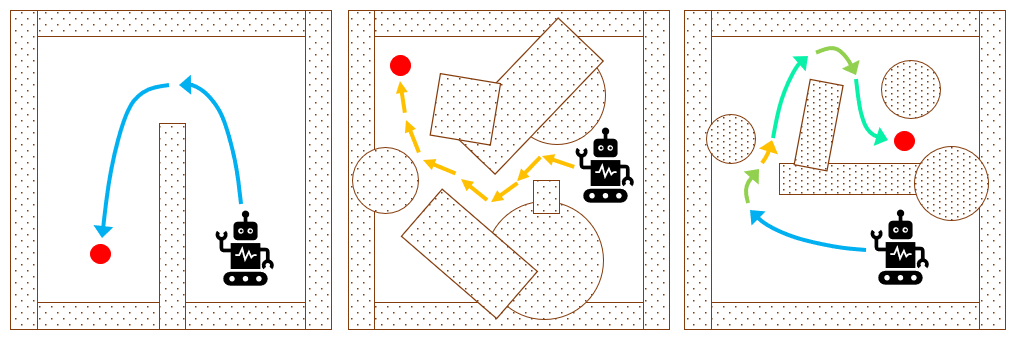}}
		\caption{Motivation: In the scenarios with different obstacle distributions, the robot has different ideal forward simulation time (which can be considered as desired action duration in RL). Therefore, we propose Adaptive Forward Simulation Time (AFST).}
		\label{fig0}
	\end{figure}

	Before presenting our method, we introduce two crucial parameters in conventional navigation methods: \textit{forward simulation time (FST)} and \emph{control interval}. 
	\textit{FST} represents the time for forward-simulating trajectories, indicating how far ahead the robot plans. 
	\textit{Control interval} is the reciprocal of the control frequency, determining the duration of executing each command. 
	Clearly, the smaller the \textit{control interval} is, the more timely the robot can react to its surroundings. 
	However, there usually exists a “sweet value range” for the choice of \textit{FST}. 
	If the \textit{FST} is too small, the robot may be trapped in local minimums due to short-term planning. 
	Conversely, a longer \textit{FST} makes it challenging for the robot to navigate in obstacle-filled environments. 
	Hence, \textit{FST} plays a critical role in the performance of classical navigation methods~\cite{dwa,teb}.
	In most DRL-based navigation methods, the \textit{FST} is set equal to the \textit{control interval} and they are both equal to a fixed-value action duration.
	As mentioned above, these approaches will be plagued by the problems associated with excessively large or small \textit{FST} values.
	Therefore, we propose the introduction of adaptive \textit{FST} in DRL navigation, as illustrated in Fig.~\ref{fig0}.
	
	Similar to the idea of adaptive \textit{FST}, there are existing works in reinforcement learning that employ dynamic action repetition, allowing the agent to calculate an action once and repeat it for multiple steps, which results in lower planning frequency and improved real-time performance~\cite{learning2repeat}. 
	Moreover, action repetition enables the transitions between distant advantageous states with a single decision~\cite{durugkar2016deep, dar}. 
	These transitions are crucial for guiding the robot to escape from local minimums.
	
	Inspired by dynamic action repetition, we propose Adaptive Forward Simulation Time (AFST), a DRL-based navigation method capable of handling complex unknown environments with various obstacle distributions by dynamically adjusting the \textit{FST}. 
	To appropriately model this planning problem with different action durations, we formulate the navigation task as a Semi-Markov Decision Process (SMDP) (SMDP)~\cite{sutton1999between}, where each action comprises three components: translational velocity, rotational velocity, and duration.
	
	To efficiently adapt the \textit{FST}, we face the following two main challenges: the larger volume of the continuous action space that needs to be sampled and the difficulty in achieving stable improvement despite diverse action durations.
	To address the first challenge, we introduce a transformation that converts our 3D action space into a 2D trajectory parameter space (TP-space)~\cite{TP}. 
	This conversion reduces the action space and improves the effectiveness of DRL training. 
	For the second challenge, we enhance the Distributed Proximal Policy Optimization (DPPO)~\cite{dppo} for the SMDP problem. 
	We achieve stable improvement by better estimation of the policy gradient for SMDP. The default policy estimator of DPPO, Generalized Advantage Estimation (GAE) ~\cite{gae}, focuses on MDP problems. 
	Therefore, we revise the discount rate in GAE to suit our SMDP problem.
	The extended GAE (EGAE) allows us to trade off variance and bias in SMDPs, and we demonstrate that under certain circumstances, EGAE can theoretically introduce no bias.
	
	We extensively evaluate our approach in both simulated and real-world scenarios to assess its effectiveness. The experimental results demonstrate that AFST efficiently accomplishes navigation tasks in unknown environments. To further validate our method, we conduct ablation experiments that highlight the effectiveness of each component. Our contributions can be summarized as follows:
	\begin{itemize}
		\item We propose the first DRL-based navigation method modeled by a SMDP with continuous action space.
		\item  We provide a transformation that converts the 3D action space to a 2D TP-space and reduces the training difficulty.
		\item We improve the DPPO algorithm for the specified SMDP problem by modifying its GAE to better estimate the policy gradient in SMDPs
	\end{itemize}

	\section{Related Work}\label{related work}	
	The navigation problem has been formulated as a SMDP
	problem before. 
	In previous work such as~\cite{mahadevan1999robust,saha2017real}, SMDP models were used. 
	However, they all use tabular reinforcement learning algorithms, which limits their state space and
	action space. Hence, their generalization and practicability
	cannot be compared with DRL-based methods.
	
	DRL-based navigation approaches can be roughly
	divided into two categories: methods with agent-level inputs
	and methods with sensor-level inputs. 
	In specific, methods with agent-level inputs~\cite{chen2017decentralized, everett2018motion} require expensive motion data of other robots (like velocities) for the states of the DRL model, and methods with sensor-level inputs~\cite{tai2017virtual,ral2018mapless,icra2020discrete,chen2020distributed,chen2021drqn} consider sensor inputs as the states of the DRL model.
	Our method follows sensor-level inputs.
	
	Many existing methods with sensor-level inputs directly map the sensor-level inputs into the robot's low-level control commands.
	Thus, only transitions between temporally adjacent states are allowed in their framework, which makes it difficult to escape the local minimums.
	A few approaches~\cite{brito2021go,chane2021goal} try to solve long-horizon tasks by selecting subgoals as actions and reaching subgoals with a fine-tuned prior policy. 
	These methods use a hierarchical architecture which is an efficient way to decompose difficult tasks.
	However, their performance heavily depends on the fixed prior policy which cannot be improved during training and requires an extensive amount of engineering effort.
	Differently, our policy is entirely learned from interactive experience with little prior knowledge.
	
	The difference between our method and normal DRL-based methods is that we introduce the time dimension to the action space. 
	This duration of executing an action has a significant impact on the exploration efficiency and the variance of value function estimation in DRL, which has been illustrated in~\cite{durugkar2016deep,dar}.

	\section{Preliminaries}\label{prelim}
	
	\subsection{Semi-Markov Decision Process.}
	
	To treat temporal abstraction as a minimal extension of the
	reinforcement learning framework, the theory of semi-Markov Decision Process (SMDP)~\cite{sutton1999between} is built. A SMDP is a tuple $(S, A, P, R, T, \gamma)$, where $S$ is the state space, $A$ is the action space, $T$ is the transition time space, and $\gamma \in(0,1]$ is the discount factor. 
	We assume the environment has transition dynamics $P\left(s^{\prime}, \tau \mid s, a\right)$ which is unknown to the agent, where $\tau$ represents the time between taking action~$a$ in observed state~$s$ and arriving in the next state~$s^{\prime}$ and can take a new action. 
	In general, we are given the reward function $r$ for the reward after observing~$s^{\prime}$.
	The goal throughout is to learn a policy maximizing long-term expected rewards $\mathbb{E}\left[\sum_{i=0}^{L-1} \gamma^{t_{i}} r_{t_{i}}\right]$
	with a time horizon $T=t_{L}$, where $r_{t_{i}}$ is the reward from action $a_{t_{i}}$, $t_0 = 0$, $t_{i}=\sum_{j=0}^{i-1} \tau_{t_{j}}$, and $\tau_{t_{j}}$ is the duration of action $a_{t_{j}}$ which is executed from $t_{j}$.

	\subsection{Formulation of Navigation in SMDP.}\label{formulation}
	In this paper, we specify the navigation task as a SMDP problem.
	Specifically, an action~$a$ is formulated as a triple $(v, \omega, d)$, where $v$ and $\omega$ denote the required translational and rotational velocities for the robot, and $d$ denotes the execution duration for both velocities (i.e. \textit{FST}). 
	A state~$s$ consists of the pose of the robot, i.e., $(x, y, \theta)$ for the robot's position $(x,y)$ and the orientation $\theta$, as well as the corresponding observation of the robot at the position.
	If there is no collision during the execution, then the robot would execute the velocities $v$ and $\omega$ for the time interval $d$, which results in the next state $s'$.
	However, the robot may collide and end an action before completing its given execution duration, which results in the transition dynamics $P\left(s^{\prime}, \tau \mid s, a\right)$, with $\tau \in [0, d]$ denoting the actual execution time.
	A policy $\pi(a\mid s)$ specifies the probability of taking action~$a$ at state $s$. The reward function $r$ encourages the navigation policy to avoid collisions and arrive at the target position as soon as possible. 
	
	\subsection{Trajectory Parameter Space with Velocities}\label{TP-space}
	To reduce the action space and increase the probability of sampling the appropriate actions, we convert our 3D action space to a 2D trajectory parameter space (TP-space)~\cite{blanco2008extending} for planning and convert the action back to 3D for actual execution. 
	
	A trajectory parameter space is a 2D space where each point corresponds to a drivable trajectory for a robot.
	The basic idea is that a sequence of robot's pose in 3D space $(x, y, \theta)$ can be charted employing 2D manifolds in the parameter space of a family of trajectories.
	
	In this paper, we consider differential-drive robots, which only move along circular trajectories due to their movement restrictions.
	Then we improve the original circular parameter trajectory generator (C-PTG)\footnote{The original C-PTG assumes that the robot moves along the path with the maximum translational velocity all the time, which ignores the kinematic constraint of rotational velocities. Our revision not only considers this constraint but also removes the assumption of translational velocity.}~\cite{blanco2008extending} and uses a 2D virtual action $(v^\TP, \omega^\TP)$ to parameterize the trajectories with a predefined fixed time scale $\tau^\TP$. We call it virtual action because $v^\TP$ and $\omega^\TP$ are virtual translational and rotational velocities in TP-space, which are not limited by the actual robot's maximum translational velocity $v_m$ and rotational velocity $\omega_{m}$.
	
	Assume the robot starts from the initial point $(0, 0)$ and executes the virtual action $(v^\TP, \omega^\TP)$ for a time interval $\tau^\TP$. 
	The path that the robot moves along is the same as the paths of the set $\Phi$ of executable actions, such that
	{\small
		\[
		\Phi = \left\{(v, \omega, d)\mid v\, d = v^\TP\, \tau^\TP \text{ and } \frac{v}{\omega}=\frac{v^\TP}{\omega^\TP} \right\}.
		\]}%
	If we want the robot to move along this path in the shortest possible time, we can convert the virtual action $(v^\TP, \omega^\TP)$ to a unique executable action $(v, \omega, d)$, i.e., 
	{\small
		\begin{align}\label{2Dto3D}
			v &= \frac{v^{\TP}}{k}, &
			\omega & = \frac{\omega^{\TP}}{k}, &
			d &= k \, \tau^\TP, 
	\end{align}}%
	where $k$ is a real number, with $k=\max(\frac{v^{\TP}}{v_{m}},\,\frac{\omega^{\TP}}{\omega_{m}})$. 
	In this way, we only need to consider the most efficient action $(v, \omega, d)$ for the navigation policy, instead of all possible actions in $\Phi$. Hence, we reduce the action space.

	\section{Extending GAE for SMDP}\label{EPG}
	
	In reinforcement learning, events (e.g., a robot collides) in the far future are weighted less than events in the immediate future. In our scheme, an action's execution duration can determine when an event will take place in the future, and thus also determine the event's weight.
	However, original generalized advantage estimation (GAE)~\cite{gae} focuses on MDPs and does not consider such impact of execution duration.
	To address this issue, we improve GAE to estimate the policy gradient in SMDPs.
	
	First, we introduce the definitions of the state value function $V^{\pi_\theta}$, the state-action value function $Q^{\pi_\theta}$, and the advantage function $A^{\pi_\theta}$ in SMDP, i.e.,
	{\small
		\begin{align*}
			V^{\pi_\theta}(s_{t_{i}}) &= \mathbb{E}_{\rho \sim p_{\theta}(\rho_{s_{t_i}})}\Big[\sum_{j=0}^{\infty} \gamma^{z_{i}^{j}} r_{t_{i+j}}\Big],\\
			Q^{\pi_\theta}(s_{t_{i}},a_{t_i}) &= \mathbb{E}_{\rho \sim p_{\theta}(\rho_{{s_{t_i}, a_{t_i}}})} \Big[\sum_{j=0}^{\infty}\gamma^{z_{i}^{j}} r_{t_{i+j}}\Big],\\
			A^{\pi_\theta}(s_{t_{i}},a_{t_i}) &= Q^{\pi_\theta}(s_{t_i}, a_{t_i}) - V^{\pi_\theta}(s_{t_i}),
	\end{align*}}%
	where $\rho_{s_{t_i}}$ (resp. $\rho_{s_{t_i}, a_{t_i}}$) denotes the episode starting from the state $s_{t_{i}}$ (resp. the state $s_{t_{i}}$ and the action $a_{t_i}$), and $z_{i}^{j}=t_{i+j}-t_{i}=\sum_{k=i}^{k=i+j-1} \tau_{t_k}$.

	An advantage estimator $\hat{A}_{\rho}$ denotes the estimation of the advantage function by considering the episode $\rho$.
	Given an approximate state value function $\hat{V}$, for each $k\geq 1$ we define the advantage function of $k$ step
	{\small
		\begin{multline*}
			\hat{A}_\rho^{(k)}(s_{t_i}, a_{t_i}) = \sum_{j=0}^{k-1} \gamma^{z_i^j} \delta_{t_{i+j}}^{\hat{V}} =\\
			-\hat{V}(s_{t_{i}})+r_{t_{i}} + \gamma^{z_i^1} r_{t_{i+1}}+\cdots+ \gamma^{z_{i}^{k-1}} r_{t_{i+k-1}} +\gamma^{z_{i}^{k}} \hat{V}(s_{t_{i+k}}),
	\end{multline*}}%
	where $z^j_i=t_{i+j} - t_i$ and
	$\delta_{t_{i+j}}^{\hat{V}}=  r_{t_{i+j}}+\gamma^{z_{i+j}^1} \hat{V}(s_{t_{i+j+1}}) -\hat{V}(s_{t_{i+j}})$.
	In particular,
	{\small
		\begin{align*}
			\hat{A}_\rho^{(1)}(s_{t_i}, a_{t_i}) &= \delta_{t_i}^{\hat{V}},\\
			\hat{A}_\rho^{(\infty)}(s_{t_i}, a_{t_i}) &= -\hat{V}(s_{t_{i}})+\sum_{j=0}^{\infty} \gamma^{z_{i}^{j}} r_{t_{i+j}}.
	\end{align*}}%
	Then we define EGAE $\hat{A}_\rho^{\EGAE}$
	{\small
		\begin{align*}
			&\hat{A}_\rho^{\EGAE}(s_{t_i}, a_{t_i})\\ 
			&= (1-\lambda) \left(\hat{A}_\rho^{(1)}(s_{t_i}, a_{t_i})+\lambda \hat{A}_\rho^{(2)}(s_{t_i}, a_{t_i})+\cdots\right) \\
			&= (1-\lambda)\left(\delta_{t_i}^{\hat{V}}+\lambda\left(\delta_{t_i}^{\hat{V}}+\gamma^{z_{1}^{i}} \delta_{t_{i+1}}^{\hat{V}}\right)+\cdots\right)\\
			&= \sum_{j=0}^{\infty} \gamma^{z_{i}^{j}} \lambda^{j} \delta_{t_{i+j}}^{\hat{V}},
	\end{align*}}%
	where $\lambda\in [0, 1]$ is a hyperparameter representing the compromise between bias and variance. Similar to GAE, the increase of $\lambda$ results in the increase of the variance and the decrease of the bias. 
	In the settings of SMDPs, we prove that EGAE can theoretically be an estimator that introduces no bias when $\lambda=1$ or $\hat{V}$ is accurate. The detailed derivation process and more research on the properties of EGAE are presented in the appendix\footnote{\url{https://github.com/YohannnChen/AFST/blob/main/appendix.pdf}. The appendix material mentioned below can also be found at this link.}.

	\section{Approach}\label{approach}
	In this section, we present details on how to apply Distributed Proximal Policy Optimization (DPPO)~\cite{dppo} to the SMDP problem with the adoption of EGAE. 
	DPPO is a distributed implementation of Proximal Policy Optimization (PPO)~\cite{schulman2017proximal}. 
	We consider DPPO in the paper as PPO and its extensions are widely applied in DRL-based robot navigation due to their simplicity, stability, and high sample efficiency~\cite{marchesini2021benchmarking, ral2018mapless, chen2020distributed, brito2021go}. 
	More importantly, it is a policy-based DRL algorithm, which can learn stochastic policies.
	Notice that, for a navigation policy that drives the robot in unknown environments, it would frequently face aliased states~\cite{silver2015lecture} due to its partial observation of the environment, while a stochastic policy can help the robot to address the problem.
	
	\subsection{Reinforcement Learning Components.}\label{sec:drl}
	
	\subsubsection{\bf \emph{State}}
	a state $s_{t_i}$ consists of the egocentric local grid map~$M_{t_i}$ and the relative target pose $g_{t_i}$ of the robot as shown in the Fig.~\ref{fig2}. 
	In our experiment, $M_{t_i}$ is generated from the data of a 2D laser scan with a 180-degree horizontal Field of View (FOV), which encodes the robot's shape and observable appearances
	of nearby obstacles. 
	We construct the egocentric local grid map $M_{t_i}$ in the same way as~\cite{chen2020distributed}.
	
	\subsubsection{\bf \emph{Action}}
\begin{wrapfigure}{r}{0.2\textwidth}
		\centering
		\includegraphics[width=0.2\textwidth]{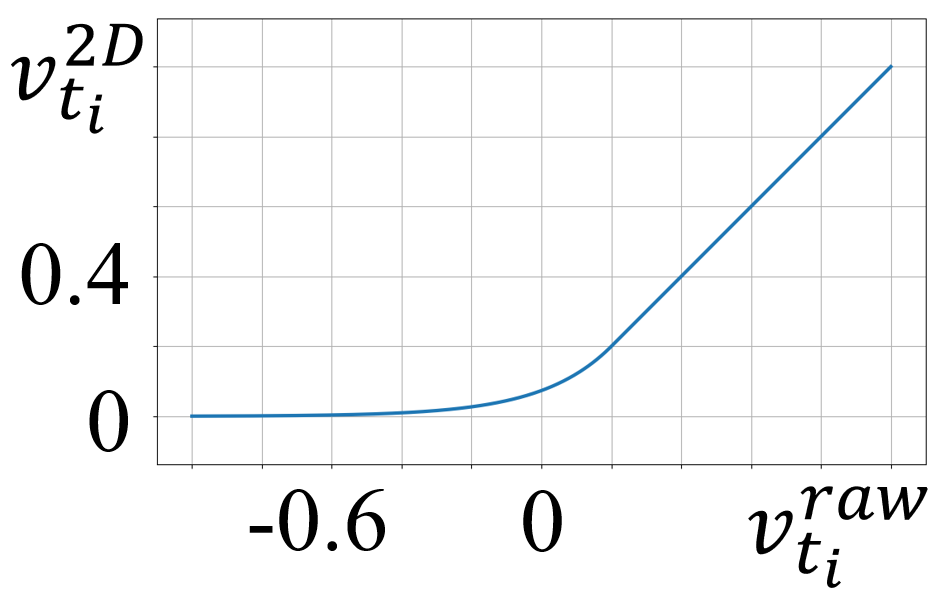}
		\caption{Our modified ELU}
		\label{fig7}
	\end{wrapfigure}

\begin{figure}[tp]
		\centerline{\includegraphics[width=1\linewidth]{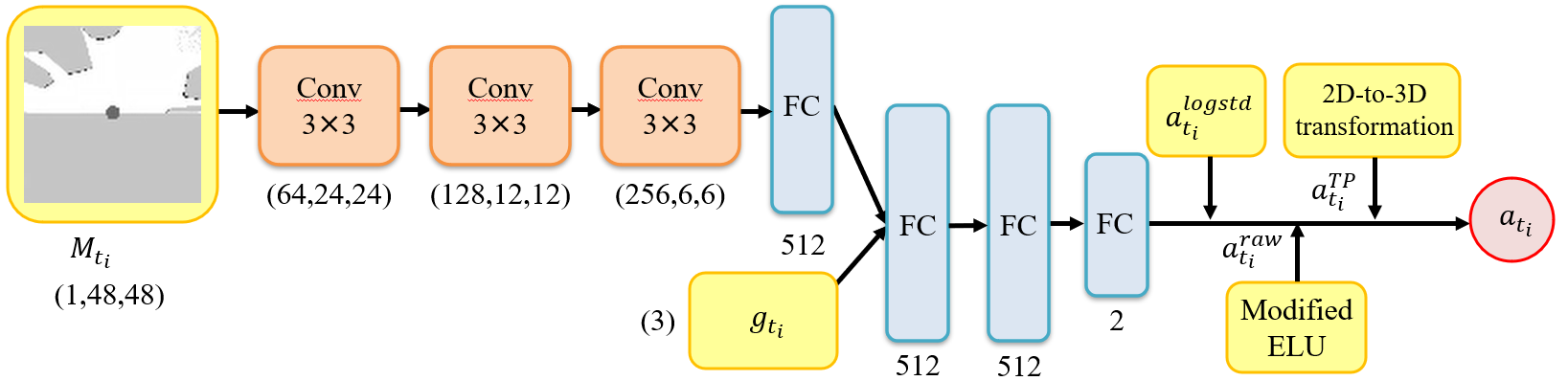}}
		\caption{We use DPPO to solve the navigation problem modeled in a SMDP. Above is the architecture of our policy network. The input of the network is $s_{t_i}$, which consists of the local grid map $M_{t_i}$ and the relative target pose $g_{t_i}$. The output of the last layer is added by a Gaussian noisy $a_{t_i}^{logstd}$ and gets the raw action $a_{t_i}^{raw}$. Then we use a specified activation function with an ELU to process it and get the 2D virtual action $a_{t_i}^{TP}$. Finally, we  convert $a_{t_i}^{TP}$ to 3D action $a_{t_i}$ for execution according to Eq.~\eqref{2Dto3D}}
		\label{fig2}
	\end{figure}

	an executable action $a_{t_i}$ for the robot is a triple $(v_{t_i}, \omega_{t_i}, d_{t_i})$.
	We adopt the conversion illustrated in Sec.~\ref{TP-space} to reduce the action space.
	Here, we introduce the processing process from 2D raw actions to 3D executable actions as shown in Fig.~\ref{fig2}. Firstly, the output of the last layer is added by a Gaussian noisy $a_{t_i}^{logstd}$ for exploration. Then, we get the raw action $a_{t_i}^{raw}$. 
	Secondly, we specify an activation function with a modified exponential linear unit (ELU)~\cite{ELU} to convert the raw action $a_{t_i}^{raw}$ to a 2D virtual action $a_{t_i}^{TP}$, where $a_{t_i}^{raw}=(v^\raw_{t_i}, \omega^\raw_{t_i})$, $a_{t_i}^{TP}=(v^{TP}_{t_i}, \omega^{TP}_{t_i})$,
	{\small
		\begin{equation}\label{activate}
			\begin{aligned} 
				v^{TP}_{t_i}&=\left\{
				\begin{array}{ll}
					0.2 e^{5 v^{\raw}_{t_i}-1} & \text { if }v^{\raw}_{t_i}<0.2, \\
					v^{\raw}_{t_i} & \text { otherwise, }
				\end{array}\right.\\
				\omega^{TP}_{t_i}&=\omega^{\raw}_{t_i}.
			\end{aligned} 
	\end{equation}}%
	We adopt the ELU as it can improve the robustness to noise when the robot needs subtle movements to pass through dense obstacles.
	As shown in Fig.~\ref{fig7}, our ELU gradually saturate to zero when the argument gets smaller. We can observe that the smaller the argument is, the slower the gradient vanishes, which preserves ELU's property of refining the granularity of 
	small values. 
	This is why the ELU has the characteristic of soft saturation and improves the robustness to noise.
	The parameters of the ELU influence the probability distribution of the different actions sampled by the initial policy. Tested by experiments, we find that the ELU's parameters in Eq.~\eqref{activate} are preferable.
	Finally, $a_{t_i}^{TP}$ can be converted to $a_{t_i}$ for execution by Eq.~\eqref{2Dto3D}.
	
	\subsubsection{\bf \emph{Reward}}
	the reward $r_{t_{i}}$ consists of four components, i.e.,
	{\small
		\begin{align*}
			r_{t_{i}} &=r^{app}_{t_{i}}+r^{arr}_{t_{i}}+r^{col}_{t_{i}}+r^{step}_{t_{i}},\\
			r^{app}_{t_{i}} &=\varepsilon_{a}\left(\left\|{l}_{t_{i-1}}-l_{g}\right\|-\left\|{l}_{t_{i}}-l_{g}\right\|\right), \\
			r^{arr}_{t_{i}} &=\left\{\begin{array}{ll}
				r_{arr}  & \text { if }\left\|{l}_{t_{i}}-l_{g}\right\|<0.3,\\
				0 & \text{ otherwise,}
			\end{array}\right.\\
			r^{coll}_{t_{i}} &=\left\{\begin{array}{ll}
				r_{col} & \text { if collision}, \\
				0 & \text { otherwise,}
			\end{array}\right.\\
			r^{step}_{t_{i}} &= -\varepsilon_{t} \tau_{t_{i}} - \varepsilon_{\tau} \tau^\TP,
	\end{align*}}%
	where $l_{t_i}$ (resp. $l_{g_{t_i}}$) denotes the location of the robot (resp. target $g_{t_i}$) at the current time $t_i$, $\tau_{t_{i}}$ denote the actual execution duration of $a_{t_i}$, $r_{arr}>0$, $r_{col} <0$, $\varepsilon_{a}>0$, $\varepsilon_{t}>0$, and $\varepsilon_{\tau}>0$ are hyperparameters.
	
	In particular, $r^{app}_{t_{i}}$ defines the penalty for departure from the target,
	$r^{arr}_{t_{i}}$ defines the reward for arrival at the target, $r_{col}$ defines the penalty for collisions, and $r^{step}$ defines a minor penalty for each action. $\varepsilon_{t}$ is set to penalize the execution time of the action, which encourages the robot to reach the goal faster.
	$\varepsilon_{\tau}$ is the coefficient of a fixed penalty for every action, which encourages the robot to reach the goal with fewer planning times.
	In our experiment, we set $r_{arr} = 500$, $r_{col} = -500$,  $\varepsilon_{a}=200$, $\varepsilon_{t}=12$, and $\varepsilon_{\tau} = 10$.
	
	\subsection{Network Architecture.}
	
	The architecture of our policy network~$\pi_\theta$ is shown in Figure~\ref{fig2}.
	Our value network has the same architecture as the policy network, except that the last layer is modified to only output the value function~$V_\phi$ without the Gaussian noise and the activation function.

	\subsection{Training Process.}
	
	We replace GAE with EGAE in DPPO. Then the objective function for the policy network $\pi_\theta$ is
	{\small
		\begin{align*}
			L^{\pi}_{\theta_{k}}(\theta) &=\mathbb{E}_{s, a \sim \theta_{k}} \Big[ \min \Big(\frac{\pi_{\theta}(a \mid s)}{\pi_{\theta_{k}}(a \mid s)} \hat{A}_\rho^\EGAE(s, a), \\
			&\hspace{3.5cm} g(\epsilon,\, \hat{A}_\rho^\EGAE (s, a))\Big)\Big], \\
			g(\epsilon, \hat{A}) &=\left\{\begin{array}{ll}
				(1+\epsilon) \hat{A} & \text{ if }\hat{A} \geqslant 0 \\
				(1-\epsilon) \hat{A} & \text{ otherwise,}
			\end{array}\right.
	\end{align*}}%
	where $\theta_{k}$ denotes the parameters of the policy at the $k$th epoch and $\epsilon$ is the clip ratio.
	
	The objective function for the value network $V_\phi$ is
	{\small
		\begin{align*}
			L^{V^{\pi}}_{\phi}(\phi) &=- \mathbb{E}_{s, \rho_{s} \sim \pi} \Big[ \Big( R(\rho_s)-V_{\phi}(s) \Big)^2 \Big],
	\end{align*}}%
	where $\rho_s$ denotes the episode starting from the state $s$ and $\phi$ denotes the parameters of the value function. $L^{\pi}_{\theta_{k}}(\theta)$ and $L^{V^\pi}_{\phi}(\phi)$ will both be estimated per iteration by all the data in the experience buffer.

	\begin{algorithm}[tp]
		\caption{Adaptive Forward Simulation Time (AFST)}
		\label{alg:1}
		\begin{algorithmic}[1]
			\STATE Initialize the policy network $\pi_{\theta}$, the value network $V_\phi$.\
			\STATE Clear the experience buffer \emph{Buffer}.\
			\FOR{\emph{epoch} $k= 1,2,...,$ } 
			\STATE \emph{//collect data in parallel}
			\FOR{\emph{step} $i= 0,1,...,T_{ep}$ }
			\STATE $a_{t_i}^{raw}=\pi_{\theta}(s_{t_i}); V_{t_i}=V_\phi(s_{t_i})$
			\STATE Transfer $a_{t_i}^{raw}$ to $a_{t_i}$ according to Eq.~\eqref{activate}, Eq.~\eqref{2Dto3D}
			\STATE $s_{t_{i+1}}, r_{t_i} = \emph{step}(a_{t_i})$
			\STATE Add $(s_{t_i}, a_{t_i}^{raw}, r_{t_i}, V_{t_i})$ to \emph{Buffer}
			\STATE $s_{t_i} \gets s_{t_{i+1}}$
			\IF{$\emph{trajectory length}>T_m$ or \emph{arrive} or \emph{collide}}
			\STATE Finish the current episode $\rho$ by $V_{t_{i+1}}=0$
			\STATE Estimate $\hat{A}_{\rho}^\EGAE(s_{t_j}, a_{t_j})$ for all $(s_{t_j}, a_{t_j})$ in $\rho$
			\STATE Add each $\hat{A}_{\rho}^\EGAE(s_{t_j},a_{t_j})$ to \emph{Buffer}
			\STATE Add each $R(\rho_{s_{t_j}})$ to \emph{Buffer}
			\STATE $s_{t_i}=\emph{reset()}$
			\ENDIF
			\ENDFOR
			\STATE Update $\pi_{\theta}$ with $L^{\pi}_{\theta_{k}}(\theta)$ and $lr_\theta$ for $E_\pi$ iterations
			\STATE Update $V_\phi$ with $L^{V^\pi}_{\phi}(\phi)$ and $lr_\phi$ for $E_v$ iterations
			\STATE Clear \emph{Buffer}
			\ENDFOR
		\end{algorithmic}
	\end{algorithm}
	
	As illustrated in Figure~\ref{fig3:subfig:a},
	we train the networks in environments that are constructed by a customized simulator based on OpenCV with Algorithm~\ref{alg:1}.
	In particular, the simulator can load an environment as a gray image, where obstacles and robots are specified as corresponding pixels in the image. 
	In DPPO, the algorithm collects experiences in a distributed setting from multiple environments where robots share the same navigation policy~$\pi_\theta$.
	
	\section{Experiments}
	
	In this section, we present the simulated and physical evaluation methods. Their results both validate our hypothesis that AFST can achieve better performance in complex unknown environments. 
	
	\subsection{Implementation Details}
	We specify two modes to implement the navigation policy:
	\begin{itemize}
		\item {\bf \emph{Mode1}}: the \emph{control interval} is equal to FST, which means executing the action as long as the algorithm plan ahead.
		\item {\bf \emph{Mode2}}: the \emph{control interval} is set to a fixed small value, which means that no matter how long the FST is planned, the robot will replan the next action after executing the current action for the fixed small interval.
	\end{itemize}
	In most cases, we use \emph{Mode1}, such as simulation training, simulation testing, and physical experiments in static scenarios. \emph{Mode2} is used in physical experiments with dynamic obstacles for better safety. We set the fixed \emph{control interval} of \emph{Mode2} to $0.1s$.
	
	In our experiments, we
	use a turtlebot2, a differential drive, a circular robot with a radius of $0.17m$, the maximum velocities of which are $v_m=0.6$ m/s and $\omega_{m}=0.9$ rad/s. The local grid maps are $6m \times 6m$. The robot is in the middle of the local map and can only perceive the information around $3m$ thereby cannot plan too far ahead.
	These settings are the same in both simulation and real-world experiments.
	
	We train the AFST with hyperparameters listed in the appendix. 
	In the experiments, we find out that almost every value of $\tau^\TP$ results in a similar success rate\footnote{Detailed evaluations can be found in the appendix.}.
	Among the values of $\tau^\TP$, $0.4s$ is slightly better.
	Both the policy and value networks are implemented in TensorFlow and trained with the Adam optimizer on a computer with i7-9900 CPU and Nvidia Titan RTX GPU. 
	It takes around 10 hours to run about 1100 epochs in DPPO when the networks converge in the training.

	\subsection{Simulated Experiments}
	\subsubsection{Training and Testing Scenarios}

	\begin{figure}[tp]
		\centering
		\subfigure[Sparse]{
			\label{fig3:subfig:a} 
			\includegraphics[width=0.16\linewidth]{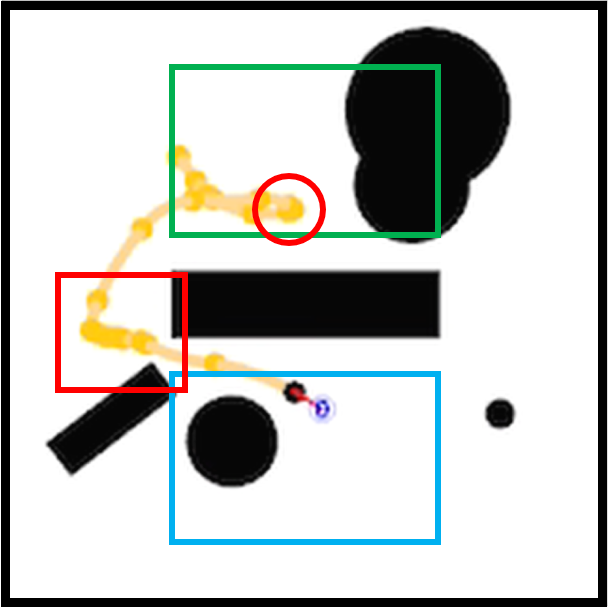}
		}
		\subfigure[Dense]{
			\label{fig3:subfig:b}
			\includegraphics[width=0.16\linewidth]{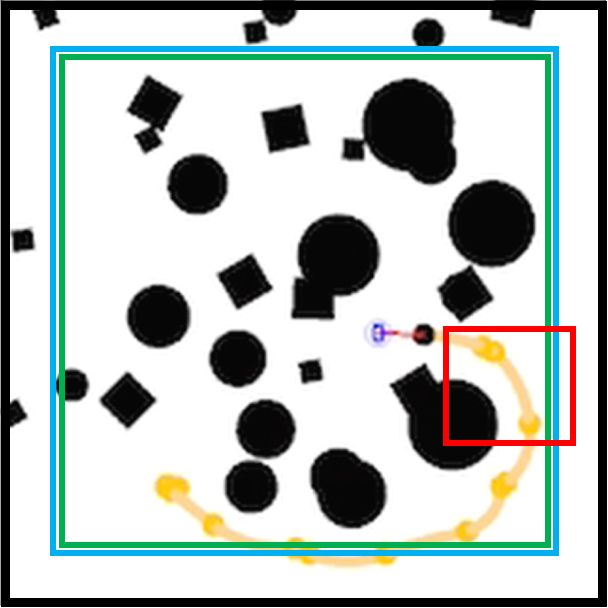}
		}
		\subfigure[Spiral]{
			\label{fig3:subfig:c} 
			\includegraphics[width=0.16\linewidth]{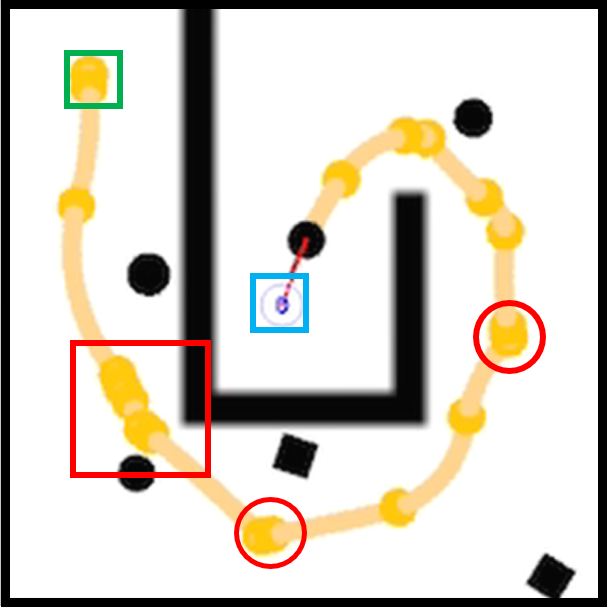}
		}
		\subfigure[Zigzag]{
			\label{fig3:subfig:d}
			\includegraphics[width=0.16\linewidth]{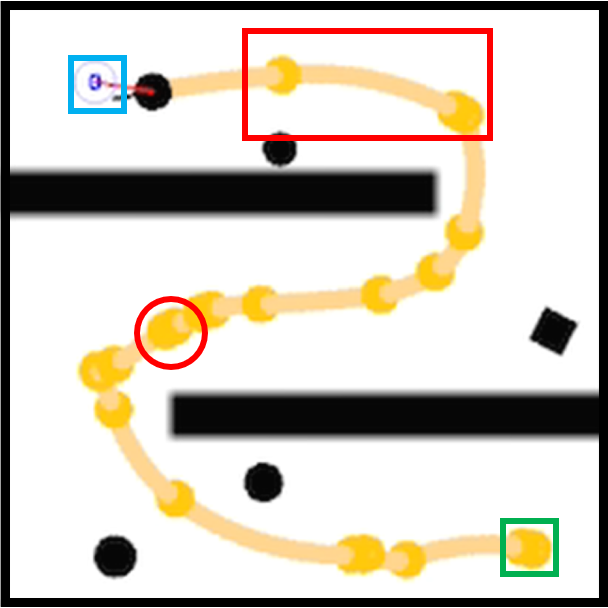}
		}
		\subfigure[Hybrid]{
			\label{fig3:subfig:e}
			\includegraphics[width=0.16\linewidth]{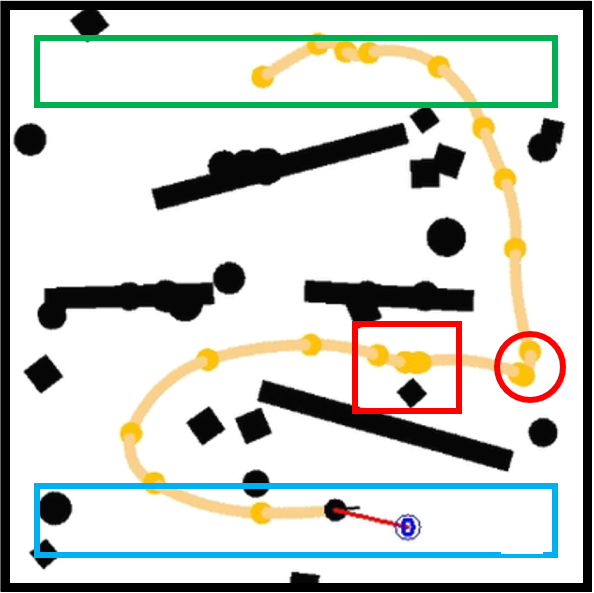}
		}
		\caption{We consider five scenarios in our experiments and their global top views are shown as above, where yellow curves denote the navigation paths of the robot, the yellow dots denote the planning moment, the green boxes denote the random 
			region of the starting position, and the blue boxes denote the random region of the target position.  
			We can see in the red boxes that our robot can adjust its forward simulation time to adapt to multiple spatial features and obstacle densities. More importantly, our robot successfully escapes the local minimums marked with red circles. }
		\label{fig3}
	\end{figure}

	Figure~\ref{fig3} illustrates five different scenarios that are considered in our experiments. 
	Please note that we only consider environments in \emph{sparse} scenario as training environments and evaluate the performance in these five scenarios:
	
	\begin{itemize}
		\item {\bf \emph{Sparse}} scenario is $10m \times 10m$ and randomly chooses locations for 6 randomly sized obstacles. The starting and the target positions of the robot are also randomly chosen. 
		So much randomness is to avoid overfitting.
		%\footnote{Specifically, the position of the large rectangle in the middle is fixed, and the positions of other obstacles are random within the scope of the whole room. The advantage of fixing the position of the large rectangle is making sure that there are obstacles between the starting and the target position of the robot, and it is more likely to form local minimum traps.}
		\item {\bf \emph{Dense}} scenario is $10m \times 10m$ and randomly choose locations for 32 small obstacles. The starting and the target positions are also randomly chosen.
		\item {\bf \emph{Spiral}} scenario is $6m \times 6m$, using a spiral map with 5 small obstacles, the starting, and the target positions are fixed.
		\item {\bf \emph{Zigzag}} scenario is $6m \times 6m$, using a zigzag map with 5 small obstacles, where the starting and target positions are fixed.
		\item {\bf \emph{Hybrid}} scenario is $10m \times 10m$, combining the characteristics of \emph{Dense} and \emph{Zigzag} scenarios, where the locations of 32 obstacles, the starting, and the target position are randomly chosen.
	\end{itemize}

	\subsubsection{Comparison Baselines and Evaluation Metrics}
	
	We compare AFST with four baseline methods, i.e., 
	\begin{itemize}
		\item {\bf \emph{DWA}}: Dynamic Window Approach~\cite{dwa}, an elegant and classical local planner,  with a fixed forward simulation time, which is set to 3s in our experiments.
		\item {\bf \emph{CAMDRL}}: Collision Avoidance via Map-Based Deep Reinforcement Learning (CAMDRL)~\cite{chen2020distributed}, a map-based robot navigation method with fixed action duration. Its inputs of the network are local maps generated by the data of the laser. The original GAE is applied to it.
		\item {\bf \emph{GO-DWA}}: Goal-Oriented Dynamic Window Approach (GO-DWA), a hierarchical planning approach that selects subgoals as actions and reaches the subgoals with DWA. We refer to the scheme and the action space of GO-MPC~\cite{brito2021go} and replace its MPC with DWA.
		\item {\bf \emph{SDDQN}}: Semi-Markov DoubleDQN~\cite{schmoll2021semi}, a DRL-based method to solve SMDP problems with discrete action space. We set the range of the translational velocity $v \in \{0, 0.3, 0.6\}$, the rotational velocity $\omega \in \{-0.9, 0, 0.9\}$, the planning duration  $d \in \{0.4, 0.8, 1.2, 1.6, 2.0\}$. 
		
	\end{itemize}
	For a fair comparison, all methods share the same laser data as the input, all DRL-based methods use the same reward function as AFST. The action duration in CAMDRL is set to a fixed value of 0.4s as~\cite{chen2020distributed} does. 
	
	We use \emph{success rate} (SR) and \emph{reach time} (RT) to evaluate the performance. The \emph{success rate} is the ratio of tests that the robot reaches its target within 200 steps without any collision. The \emph{reach time} is the average time taken by the robot to reach the target. The evaluation results of trajectory length are presented in the appendix.
	
	\begin{table}[bp]
		\centering
		\setlength\tabcolsep{3.5 pt}
		\caption{Evaluation results of different methods}
		\begin{tabular}{rlrrrrrr}
			\toprule
			\multicolumn{1}{l}{Metric} & Method & \multicolumn{2}{l}{\#scenario} &       &       &       & \multirow{2}[4]{*}{Average } \\
			\cmidrule{3-7}          &       & Sparse & Dense & Spiral & Zigzag & Hybrid &  \\
			\midrule
			\multicolumn{1}{l}{Success} & DWA   & 0.616 & 0.618 & 0     & 0     & 0.002 & 0.248 \\
			\multicolumn{1}{l}{rate} & CAMDRL & 0.782 & 0.760 & 0     & 0     & 0.542 & 0.418 \\
			\multicolumn{1}{l}{(SR)}& GO-DWA & 0.886 & 0.700 & 0.610 & 0.082 & 0.206 & 0.498 \\
			& SDDQN & 0.868 & 0.854 & 0.826 & 0.078 & 0.624 & 0.650 \\
			& AFST  & \textbf{0.946} & \textbf{0.910} & \textbf{1.00} & \textbf{0.904} & \textbf{0.778} & \textbf{0.908} \\
			\midrule
			\multicolumn{1}{l}{Reach} & DWA   & 39.8  & 20.4  & /     & /     & 37.6  & / \\
			\multicolumn{1}{l}{time} & CAMDRL & 19.2  & 11.2  & /     & /     & \textbf{28.3} & / \\
			\multicolumn{1}{l}{(RT)}& GO-DWA & 35.0  & 22.5  & 31.0  & \textbf{23.5} & 38.0  & 30.0 \\
			& SDDQN & 26.0  & 14.6  & 30.9  & 127   & 46.3  & 49.0 \\
			& AFST  & \textbf{16.9} & \textbf{9.61} & \textbf{23.1} & 39.5  & 34.2  & \textbf{24.7} \\
			\bottomrule
		\end{tabular}%
		\label{table2}%
	\end{table}%
	
	\begin{table}[bp]
		\centering
		\setlength\tabcolsep{3.5 pt}
		\caption{Results of an ablation study}
		\begin{tabular}{rlrrrrrr}
			\toprule
			\multicolumn{1}{l}{Metric} & Method & \multicolumn{2}{l}{\#scenario} &       &       &       & \multirow{2}[4]{*}{Average } \\
			\cmidrule{3-7}          &       & Sparse & Dense & Spiral & Zigzag & Hybrid &  \\
			\midrule
			\multicolumn{1}{l}{Success} & -ELU  & 0.936 & 0.840 & 0.990 & 0.532 & 0.556 & 0.771 \\
			\multicolumn{1}{l}{rate} & -2D-to-3D & 0.860 & 0.850 & 0.104 & 0     & 0.588 & 0.481 \\
			\multicolumn{1}{l}{(SR)}& -EGAE & 0.760 & 0.646 & 0.228 & 0     & 0.436 & 0.414 \\
			& -SMDP & 0.944 & 0.906 & \textbf{1.00} & 0.804 & 0.686 & 0.868 \\
			& AFST  & \textbf{0.946} & \textbf{0.910} & \textbf{1.00} & \textbf{0.904} & \textbf{0.778} & \textbf{0.908} \\
			\midrule
			\multicolumn{1}{l}{Reach} & -ELU  & 17.2  & \textbf{9.03} & 27.0  & 51.1  & \textbf{26.1} & 26.1 \\
			\multicolumn{1}{l}{time} & -2D-to-3D & 17.4 & 9.78  & 79.4  & /     & 34.4  & / \\
			\multicolumn{1}{l}{(RT)}& -EGAE & 17.5  & 9.16  & 33.4  & /     & 31.2  & / \\
			& -SMDP & 17.0  & 11.0  & 23.7  & \textbf{37.5} & 27.9  & \textbf{23.4} \\
			& AFST  & \textbf{16.9}  & 9.61  & \textbf{23.1} & 39.5  & 34.2  & 24.7 \\
			\bottomrule
		\end{tabular}%
		\label{table3}%
	\end{table}%

	\subsubsection{Testing Results}
	
	Table~\ref{table2} summarizes the test results using \emph{Mode1}. In every scenario, we performed 500 tests for each model. We detail our observations below:
	
	{\bf \emph{The performance of DWA}}:
	DWA has inherent safety rules in that the robot is only allowed to take collision-free actions (assuming obstacles are static).
	However, DWA cannot achieve $100\%$ SR as the robot may get trapped in the local minimum. This problem causes the SR of DWA to be decreased to zero in some scenarios and its RT to be increased in others.

	{\bf \emph{The performance of CAMDRL}}:
	As a method with fixed action duration, CAMDRL has a tolerable SR in \emph{sparse} and \emph{dense} scenarios as local minima areas don't occur very often with randomly located obstacles. However, it is entirely trapped in the local minimums of \emph{spiral} and \emph{zigzag} scenarios, the detailed analysis of this phenomenon is presented in the appendix. This phenomenon is mitigated in \emph{hybrid} scenario as the sizes of their obstacles are smaller than the ones in \emph{zigzag} scenario.
	
	{\bf \emph{The performance of GO-DWA}}:
	GO-DWA achieves a high SR in \emph{sparse} scenario. However, we find out that it is difficult for GO-DWA to select an accurate local goal between dense obstacles. In some cases, these local goals will overlap with obstacles, which leads to the limited generalization of GO-DWA.
	
	{\bf \emph{The performance of SDDQN}}:
	SDDQN's performance is the closest to AFST, which indicates the advantage of using SMDP models. But its deterministic policy is not as good at handling aliased states as stochastic policy as mentioned in Sec.~\ref{approach}. Besides, SDDQN's discrete action space limits its generalization. 
	Therefore, SDDQN has a lower SR and longer RT than AFST in testing scenarios.
	
	{\bf \emph{The performance of AFST}}:
	AFST achieves the highest SR in all testing scenarios. The high SR and the trajectories in Fig.~\ref{fig3} prove it can adjust the FST to a proper small value when running through the narrow gaps (Fig.~\ref{fig3:subfig:a}), and can adjust the FST to a proper big value when escaping the local minimums(Fig.~\ref{fig3:subfig:d}). Although the SR of AFST in the training scenario is not much higher than that of other methods, its SR in unseen testing scenarios is distinctly higher, which indicates its excellent generalization. Moreover, AFST has the shortest mean RT as it always moves with the maximum speed according to Eq.~\eqref{2Dto3D} and can escape from local minimums with fewer failed attempts.

	\subsubsection{Ablation Studies}
	
	We also conduct ablation studies to investigate the utility of several modifications we propose in our scheme. Specifically,
	`-ELU' means replacing the modified ELU with the ReLu~\cite{relu}, `-2D-to-3D' means directly outputting 3D actions without the 2D-to-3D conversion, `-EGAE' means replacing EGAE with TD(0) in the settings of the SMDP, 
	and `-SMDP' means formulating the problem as a lifted MDP\footnote{We can also model the problem with different action duration as a MDP. Because the wall time between successive time steps is not forced to be uniform in MDPs. The fixed discount rate $\gamma$ in `-SMDP' is $0.99$.} and using GAE to evaluate the advantage function.	
	The results are shown in Table~\ref{table3}. In general, we find the above modifications nearly all lead to performance improvements.

	\begin{table}[bp]
		\centering
		\caption{Performance of methods in real-world scenarios}
		\begin{tabular}{llrrrr}
			\toprule
			Metrics & Method & \multicolumn{3}{l}{\#scenario} &  \\
			\cmidrule{3-6}          &       & \multicolumn{1}{l}{Scatter1} & \multicolumn{1}{l}{Scatter2} & \multicolumn{1}{l}{Spiral} & \multicolumn{1}{l}{Zigzag} \\
			\midrule
			Success & CAMDRL & \textbf{1.0} & 0.9   & 0     & 0 \\
			rate  & AFST  & \textbf{1.0} & \textbf{1.0} & \textbf{1.0} & \textbf{1.0} \\
			\midrule
			Planning & CAMDRL & 9.16  & 10.42 & /     & / \\
			time  & AFST  & \textbf{0.62} & \textbf{1.98} & \textbf{4.14} & \textbf{1.24} \\
			\midrule
			Reach & CAMDRL & 13.23 & 27.16 & /     & / \\
			time  & AFST  & \textbf{13.18} & \textbf{25.21} & \textbf{24.06} & \textbf{28.17} \\
			\bottomrule
		\end{tabular}%
		\label{table4}%
	\end{table}%

	ELU has the characteristic of soft saturation, which improves the robustness to noise. Therefore, `-ELU' is less robust for long-distance navigation tasks and can almost only achieve short-distance tasks. This is the reason why `-ELU' has lower SR but shorter RT compared with AFST in \emph{Dense} and \emph{Hybrid} scenarios.
	Without the 2D-to-3D conversion, the action space would be much larger, which makes it much harder to sample proper actions and learn successful policies. 
	Hence, the SR of `-2D-to-3D' is low, indicating the conversion's effectiveness.
	TD(0) estimates the advantage function with a low variance but high bias. Thus the performance of ‘-EGAE’ is poor.
	For the detailed implementation, the only difference between `-SMDP' and AFST is the power of the discount rate. The experimental results show that `-SMDP' achieves similar performance to AFST. In some scenarios, it is hard to distinguish their effectiveness.
	However, MDPs can hardly evaluate the value function on an equitable time scale as SMDPs. The rewards gained from different execution duration will be multiplied by the same discount rate, encouraging long-duration actions. 
	Then, the policy of `-SMDP' is a little more aggressive than the one of AFST. Then, the average SR of `-SMDP' is lower than AFST and the average RT is shorter.
	The results of the ablation study show that each of our components plays a considerable role in the algorithm.

	\begin{figure}[tp]
		\centerline{\includegraphics[width=1.0\linewidth]{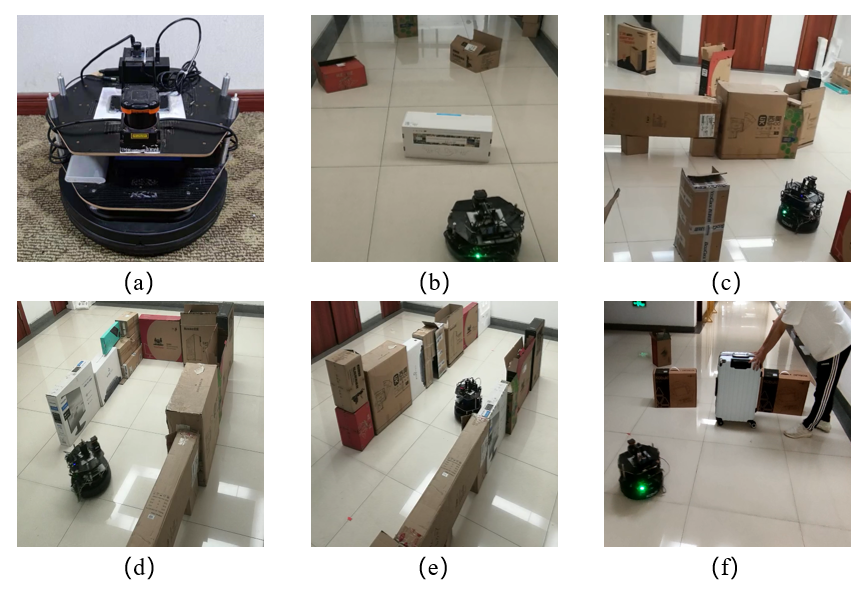}}
		\caption{We conduct real-world experiments on a TurtleBot 2 with a Kobuki base using a Hokuyo UTM-30LX 2D LiDAR and an NVIDIA Jetson TX2 (shown in (a)). We consider four static scenarios: Scattered1 (b), Scattered2 (c), Spiral (d), Zigzag (e), and a dynamic scenario (f) (with dynamic obstacles and pedestrians). }
		\label{fig5}
	\end{figure}

	\subsection{Physical Experiments}
	
	To further verify the generalization and effectiveness of our learning policy, we deployed the trained model of AFST and CAMDRL to an actual differential-drive robot in the real world as illustrated in Figure~\ref{fig5}.

	We placed paper boxes as obstacles to set up the test environments.
	To make a fair comparison, we conducted quantitative experiments in the four static scenarios and qualitative experiments in the dynamic scenario.
	In the static scenarios, we implemented AFST using \emph{Mode1}.
	For each method, we conducted 10 tests in every static scenario.
	As shown in Table~\ref{table4}, AFST can reach the target with less planning time than CAMDRL as AFST usually plans for a longer execution duration per step and AFST's network architecture is more lightweight. Moreover, CAMDRL cannot get out of the local minimums in some difficult scenarios. 
	Hence, AFST can not only tackle more difficult navigation tasks in unknown scenarios but also has the potential to navigate faster with less computational overhead.
	
	In addition, we also set up environments with dynamic obstacles and pedestrians. 
	In the \emph{dynamic} scenario, we use \emph{Mode2} to implement AFST.
	The demonstration video on both simulation and real-world experiments can be found at \url{https://youtu.be/pgP5BQHFum4}.

	\section{Discussions}\label{conc}
	\subsection{Limitations and Future Works}
	\subsubsection{\bf \emph{Dynamic collision avoidance}}
	In this paper, AFST only uses one frame of the local map for collision avoidance thereby cannot extract any dynamic information, which makes it inefficient for AFST to avoid dynamic obstacles. Besides, it is hard for the robot to react to dynamic obstacles in a timely manner only training with \emph{Mode1}. We will further develop the potential of \emph{Mode2} in dynamic scenarios.
	\subsubsection{\bf \emph{Robustness and safety}}
	Similar to many learning-based navigation methods, AFST is not as robust as optimization-based methods. The inherent exploration capabilities of AFST help it get out of local minimums, but it also increases the uncertainty of obstacle avoidance. Therefore, AFST should be tested in more complex scenarios and add more safety constraints in the future.
	\subsubsection{\bf \emph{Kinematical constraints and smoothness}}
	The transformation in TP-space assumes that the robot would move with the maximum translational velocity or rotate with the maximum rotational velocity in the environment, which is a common simplification for low-speed robots. 
	However, this assumption is unrealistic for high-speed robots. 
	One of our future works is to extend AFST for considering the Kinedynamics constraints of the robot and the smoothness of its trajectories.
	
	\subsection{Conclusion}
	In this paper, we focus on robot navigation in unknown environments. 
	To make the robot adapt to more diverse obstacle distribution,
	we propose AFST, which is the first DRL-based navigation method modeled by a SMDP with continuous action space. 
	We show that AFST outperforms several baseline schemes under multiple scenarios both in simulation and the real world.
	We also conduct ablation experiments to indicate the effectiveness of our components.

	\bibliographystyle{IEEEtran}
	\bibliography{IEEEabrv,my}

\begin{thebibliography}{10}
\providecommand{\url}[1]{#1}
\csname url@samestyle\endcsname
\providecommand{\newblock}{\relax}
\providecommand{\bibinfo}[2]{#2}
\providecommand{\BIBentrySTDinterwordspacing}{\spaceskip=0pt\relax}
\providecommand{\BIBentryALTinterwordstretchfactor}{4}
\providecommand{\BIBentryALTinterwordspacing}{\spaceskip=\fontdimen2\font plus
\BIBentryALTinterwordstretchfactor\fontdimen3\font minus
  \fontdimen4\font\relax}
\providecommand{\BIBforeignlanguage}[2]{{%
\expandafter\ifx\csname l@#1\endcsname\relax
\typeout{** WARNING: IEEEtran.bst: No hyphenation pattern has been}%
\typeout{** loaded for the language `#1'. Using the pattern for}%
\typeout{** the default language instead.}%
\else
\language=\csname l@#1\endcsname
\fi
#2}}
\providecommand{\BIBdecl}{\relax}
\BIBdecl

\bibitem{lavalle2006planning}
S.~M. LaValle, \emph{Planning algorithms}.\hskip 1em plus 0.5em minus
  0.4em\relax Cambridge university press, 2006.

\bibitem{niroui2019deep}
F.~Niroui, K.~Zhang, Z.~Kashino, and G.~Nejat, ``Deep reinforcement learning
  robot for search and rescue applications: Exploration in unknown cluttered
  environments,'' \emph{IEEE Robotics and Automation Letters}, vol.~4, no.~2,
  pp. 610--617, 2019.

\bibitem{marchesini2021benchmarking}
E.~Marchesini, D.~Corsi, and A.~Farinelli, ``Benchmarking safe deep
  reinforcement learning in aquatic navigation,'' in \emph{Proceedings of the
  34th International Conference on Intelligent Robots and Systems
  (IROS-2021)}.\hskip 1em plus 0.5em minus 0.4em\relax IEEE, 2021, pp.
  5590--5595.

\bibitem{Astar}
B.~Bonet and H.~Geffner, ``Planning as heuristic search,'' \emph{Artificial
  Intelligence}, vol. 129, no. 1-2, pp. 5--33, 2001.

\bibitem{rrt}
J.~Nasir, F.~Islam, U.~Malik, Y.~Ayaz, O.~Hasan, M.~Khan, and M.~S. Muhammad,
  ``Rrt*-smart: A rapid convergence implementation of rrt,''
  \emph{International Journal of Advanced Robotic Systems}, vol.~10, no.~7, p.
  299, 2013.

\bibitem{tai2017virtual}
L.~Tai, G.~Paolo, and M.~Liu, ``Virtual-to-real deep reinforcement learning:
  Continuous control of mobile robots for mapless navigation,'' in
  \emph{Proceedings of the 30th International Conference on Intelligent Robots
  and Systems (IROS-2017)}.\hskip 1em plus 0.5em minus 0.4em\relax IEEE, 2017,
  pp. 31--36.

\bibitem{ral2018mapless}
M.~Pfeiffer, S.~Shukla, M.~Turchetta, C.~Cadena, A.~Krause, R.~Siegwart, and
  J.~Nieto, ``Reinforced imitation: Sample efficient deep reinforcement
  learning for mapless navigation by leveraging prior demonstrations,''
  \emph{IEEE Robotics and Automation Letters}, vol.~3, no.~4, pp. 4423--4430,
  2018.

\bibitem{icra2020discrete}
E.~Marchesini and A.~Farinelli, ``Discrete deep reinforcement learning for
  mapless navigation,'' in \emph{Proceedings of the 37th International
  Conference on Robotics and Automation (ICRA-2020)}.\hskip 1em plus 0.5em
  minus 0.4em\relax IEEE, 2020, pp. 10\,688--10\,694.

\bibitem{lmp}
M.~Wang and J.~N. Liu, ``Fuzzy logic-based real-time robot navigation in
  unknown environment with dead ends,'' \emph{Robotics and autonomous systems},
  vol.~56, no.~7, pp. 625--643, 2008.

\bibitem{dwa}
D.~Fox, W.~Burgard, and S.~Thrun, ``The dynamic window approach to collision
  avoidance,'' \emph{IEEE Robotics \& Automation Magazine}, vol.~4, no.~1, pp.
  23--33, 1997.

\bibitem{teb}
M.~Keller, F.~Hoffmann, C.~Hass, T.~Bertram, and A.~Seewald, ``Planning of
  optimal collision avoidance trajectories with timed elastic bands,''
  \emph{IFAC Proceedings Volumes}, vol.~47, no.~3, pp. 9822--9827, 2014.

\bibitem{learning2repeat}
S.~Sharma, A.~Srinivas, and B.~Ravindran, ``Learning to repeat: Fine grained
  action repetition for deep reinforcement learning,'' 2017.

\bibitem{durugkar2016deep}
I.~P. Durugkar, C.~Rosenbaum, S.~Dernbach, and S.~Mahadevan, ``Deep
  reinforcement learning with macro-actions,'' \emph{arXiv preprint
  arXiv:1606.04615}, 2016.

\bibitem{dar}
A.~Lakshminarayanan, S.~Sharma, and B.~Ravindran, ``Dynamic action repetition
  for deep reinforcement learning,'' in \emph{Proceedings of the 31th AAAI
  Conference on Artificial Intelligence (AAAI-2017)}, vol.~31, no.~1, 2017.

\bibitem{sutton1999between}
R.~S. Sutton, D.~Precup, and S.~Singh, ``Between mdps and semi-mdps: A
  framework for temporal abstraction in reinforcement learning,''
  \emph{Artificial intelligence}, vol. 112, no. 1-2, pp. 181--211, 1999.

\bibitem{TP}
J.-L. Blanco, J.~Gonzalez, and J.-A. Fern{\'a}ndez-Madrigal, ``The trajectory
  parameter space (tp-space): a new space representation for non-holonomic
  mobile robot reactive navigation,'' in \emph{2006 IEEE/RSJ International
  Conference on Intelligent Robots and Systems}.\hskip 1em plus 0.5em minus
  0.4em\relax IEEE, 2006, pp. 1195--1200.

\bibitem{dppo}
N.~Heess, D.~TB, S.~Sriram, J.~Lemmon, J.~Merel, G.~Wayne, Y.~Tassa, T.~Erez,
  Z.~Wang, S.~M.~A. Eslami, M.~A. Riedmiller, and D.~Silver, ``Emergence of
  locomotion behaviours in rich environments,'' \emph{arXiv preprint
  arXiv:1707.02286}, 2017.

\bibitem{gae}
J.~Schulman, P.~Moritz, S.~Levine, M.~Jordan, and P.~Abbeel, ``High-dimensional
  continuous control using generalized advantage estimation,'' in
  \emph{Proceedings of the 4th International Conference on Learning
  Representations (ICLR-2016)}, 2016.

\bibitem{mahadevan1999robust}
S.~Mahadevan and N.~Khaleeli, ``Robust mobile robot navigation using
  partially-observable semi-markov decision processes,'' \emph{Internal
  report}, 1999.

\bibitem{saha2017real}
O.~Saha and P.~Dasgupta, ``Real-time robot path planning around complex
  obstacle patterns through learning and transferring options,'' in
  \emph{Proceedings of the 17th International Conference on Autonomous Robot
  Systems and Competitions (ICARSC-2017)}.\hskip 1em plus 0.5em minus
  0.4em\relax IEEE, 2017, pp. 278--283.

\bibitem{chen2017decentralized}
Y.~F. Chen, M.~Liu, M.~Everett, and J.~P. How, ``Decentralized
  non-communicating multiagent collision avoidance with deep reinforcement
  learning,'' in \emph{Proceedings of the 34th International Conference on
  Robotics and Automation (ICRA-2017)}.\hskip 1em plus 0.5em minus 0.4em\relax
  IEEE, 2017, pp. 285--292.

\bibitem{everett2018motion}
M.~Everett, Y.~F. Chen, and J.~P. How, ``Motion planning among dynamic,
  decision-making agents with deep reinforcement learning,'' in
  \emph{Proceedings of the 31th International Conference on Intelligent Robots
  and Systems (IROS-2018)}.\hskip 1em plus 0.5em minus 0.4em\relax IEEE, 2018,
  pp. 3052--3059.

\bibitem{chen2020distributed}
G.~Chen, S.~Yao, J.~Ma, L.~Pan, Y.~Chen, P.~Xu, J.~Ji, and X.~Chen,
  ``Distributed non-communicating multi-robot collision avoidance via map-based
  deep reinforcement learning,'' \emph{Sensors}, vol.~20, no.~17, p. 4836,
  2020.

\bibitem{chen2021drqn}
Y.~Chen, G.~Chen, L.~Pan, J.~Ma, Y.~Zhang, Y.~Zhang, and J.~Ji, ``Drqn-based 3d
  obstacle avoidance with a limited field of view,'' in \emph{Proceedings of
  the 34th International Conference on Intelligent Robots and Systems
  (IROS-2021)}.\hskip 1em plus 0.5em minus 0.4em\relax IEEE, 2021, pp.
  8137--8143.

\bibitem{brito2021go}
B.~Brito, M.~Everett, J.~P. How, and J.~Alonso-Mora, ``Where to go next:
  learning a subgoal recommendation policy for navigation in dynamic
  environments,'' \emph{IEEE Robotics and Automation Letters}, vol.~6, no.~3,
  pp. 4616--4623, 2021.

\bibitem{chane2021goal}
E.~Chane-Sane, C.~Schmid, and I.~Laptev, ``Goal-conditioned reinforcement
  learning with imagined subgoals,'' in \emph{Proceedings of the 38th
  International Conference on Machine Learning (ICML-2021)}.\hskip 1em plus
  0.5em minus 0.4em\relax PMLR, 2021, pp. 1430--1440.

\bibitem{blanco2008extending}
J.-L. Blanco, J.~Gonz{\'a}lez, and J.-A. Fern{\'a}ndez-Madrigal, ``Extending
  obstacle avoidance methods through multiple parameter-space
  transformations,'' \emph{Autonomous Robots}, vol.~24, no.~1, pp. 29--48,
  2008.

\bibitem{schulman2017proximal}
J.~Schulman, F.~Wolski, P.~Dhariwal, A.~Radford, and O.~Klimov, ``Proximal
  policy optimization algorithms,'' \emph{arXiv preprint arXiv:1707.06347},
  2017.

\bibitem{silver2015lecture}
D.~Silver, ``Lecture 7: Policy gradient,'' \emph{UCL Course on RL}, 2015.

\bibitem{ELU}
D.-A. Clevert, T.~Unterthiner, and S.~Hochreiter, ``Fast and accurate deep
  network learning by exponential linear units ({ELU}s),'' 2016.

\bibitem{schmoll2021semi}
S.~Schmoll and M.~Schubert, ``Semi-markov reinforcement learning for stochastic
  resource collection,'' in \emph{Proceedings of the 30th International
  Conference on International Joint Conferences on Artificial Intelligence
  (IJCAI-2021)}, 2021, pp. 3349--3355.

\bibitem{relu}
V.~Nair and G.~E. Hinton, ``Rectified linear units improve restricted boltzmann
  machines,'' in \emph{Proceedings of the 27th International Conference on
  Machine Learning (ICML-2010)}.

\end{thebibliography}
	
	\addtolength{\textheight}{-12cm}   % This command serves to balance the column lengths
\end{document}